\documentclass[conference]{IEEEtran}

\IEEEoverridecommandlockouts
\usepackage{cite}

\usepackage{amsmath,amssymb,amsfonts}
\usepackage{algorithmic}
\usepackage{url}
\usepackage{gensymb} 
\usepackage{enumitem}
\usepackage{graphicx}
\usepackage{textcomp}
\usepackage{xcolor}
\def\BibTeX{{\rm B\kern-.05em{\sc i\kern-.025em b}\kern-.08em
    T\kern-.1667em\lower.7ex\hbox{E}\kern-.125emX}}
\begin{document}

\title{{MarkMatch}: Same-Hand Stuffing Detection
}


\author{
    \IEEEauthorblockN{
        Fei Zhao\IEEEauthorrefmark{1},
        Runlin Zhang\IEEEauthorrefmark{2},
        Chengcui Zhang\IEEEauthorrefmark{1},
        Nitesh Saxena\IEEEauthorrefmark{3}
    }
    \IEEEauthorblockA{
        \IEEEauthorrefmark{1}Department of Computer Science,
        The University of Alabama at Birmingham,
        Birmingham, USA
    }
    \IEEEauthorblockA{
        \IEEEauthorrefmark{2}Department of Computer Science,
        University of Waterloo, Waterloo, Canada
    }
    \IEEEauthorblockA{
        \IEEEauthorrefmark{3}Department of Computer Science and Engineering,
        Texas A\&M University,
        College Station, USA \\
           larry5@uab.edu,
           r496zhan@uwaterloo.ca,
        czhang02@uab.edu,
        nsaxena@tamu.edu
    }
}

\maketitle

\begin{abstract}

We present \textit{MarkMatch}, a retrieval system for detecting whether two paper ballot marks were filled by the same hand. Unlike the previous SOTA method \textit{BubbleSig}, which used binary classification on isolated mark pairs, \textit{MarkMatch} ranks stylistic similarity between a query mark and a mark in the database using contrastive learning. Our model is trained with a dense batch similarity matrix and a dual loss objective. Each sample is contrasted against many negatives within each batch, enabling the model to learn subtle handwriting difference and improve generalization under handwriting variation and visual noise, while diagonal supervision reinforces high confidence on true matches. The model achieves an F1 score of 0.943, surpassing \textit{BubbleSig}'s best performance. \textit{MarkMatch} also integrates Segment Anything Model for flexible mark extraction via box- or point-based prompts. The system offers election auditors a practical tool for visual, non-biometric investigation of suspicious ballots.

\end{abstract}

\begin{IEEEkeywords}
Ballot Stuffing Detection, Deep Learning
\end{IEEEkeywords}

\section{Introduction}
\label{sec:intro}

Hand-marked paper ballots are a transparent and verifiable choice in many elections. However, this format remains vulnerable to same-hand stuffing attacks, where multiple marks are filled by the same individual to manipulate outcomes. For example, Barry Morphew cast a ballot for his missing wife \cite{Barry}, raising concerns about election integrity. Detecting such same-hand mark patterns is challenging, especially when the marks are small, sparse, and visually subtle.

\textit{BubbleSig}~\cite{zhao2024bubblesig}, our previous work and also the first method in this field, used a Siamese neural network to compare an isolated pair of ballot marks and predict whether they were drawn by the same person. However, the training is based solely on a single mark pair, without considering alternative candidates or broader context. This makes the network susceptible to intra-writer variability and prone to false positives in ambiguous cases where a mark resembles multiple drawing styles. Moreover, its binary classification setup optimized the comparison of a mark pair in isolation, focusing only on pairwise match probability without considering how a given mark relates to other marks. As a result, the model lacks the ability to learn relative similarity across multiple samples and therefore may struggle to find the correct match among visually similar alternatives.

To overcome these limitations, we propose \textit{MarkMatch}, taking a query mark and retrieving the most similar marks from a mark candidate pool using contrastive learning. Our contrastive model is trained with a dense similarity matrix and a dual loss objective. Each sample is contrasted against all others (row-wise and column-wise) in the batch, allowing the model to capture fine-grained handwriting variation while enforcing high confidence on true matches through diagonal supervision (see Section \ref{sec:model}). This design improves robustness to handwriting variability, ambiguous strokes, and visual noise, yielding more reliable decisions in real-world ballot settings. 

In addition, \textit{MarkMatch} integrates the Segment Anything Model (SAM) \cite{kirillov2023segment} to support prompt-based mark segmentation using bounding boxes or point clicks. Unlike the method in~\cite{zhao2023ballot}, which relies on computationally intensive alignment with blank ballot templates, SAM enables efficient, direct extraction of mark segments from marked ballots alone. This allows election officials to easily extract a query mark segment and retrieve the top-$k$ most similar instances from a mark database, offering a practical, non-biometric tool for investigating suspicious same-hand stuffing patterns.

    \vspace{-0.3cm}
  \begin{figure}[ht]
    \centering

    \includegraphics[width=0.45\textwidth]{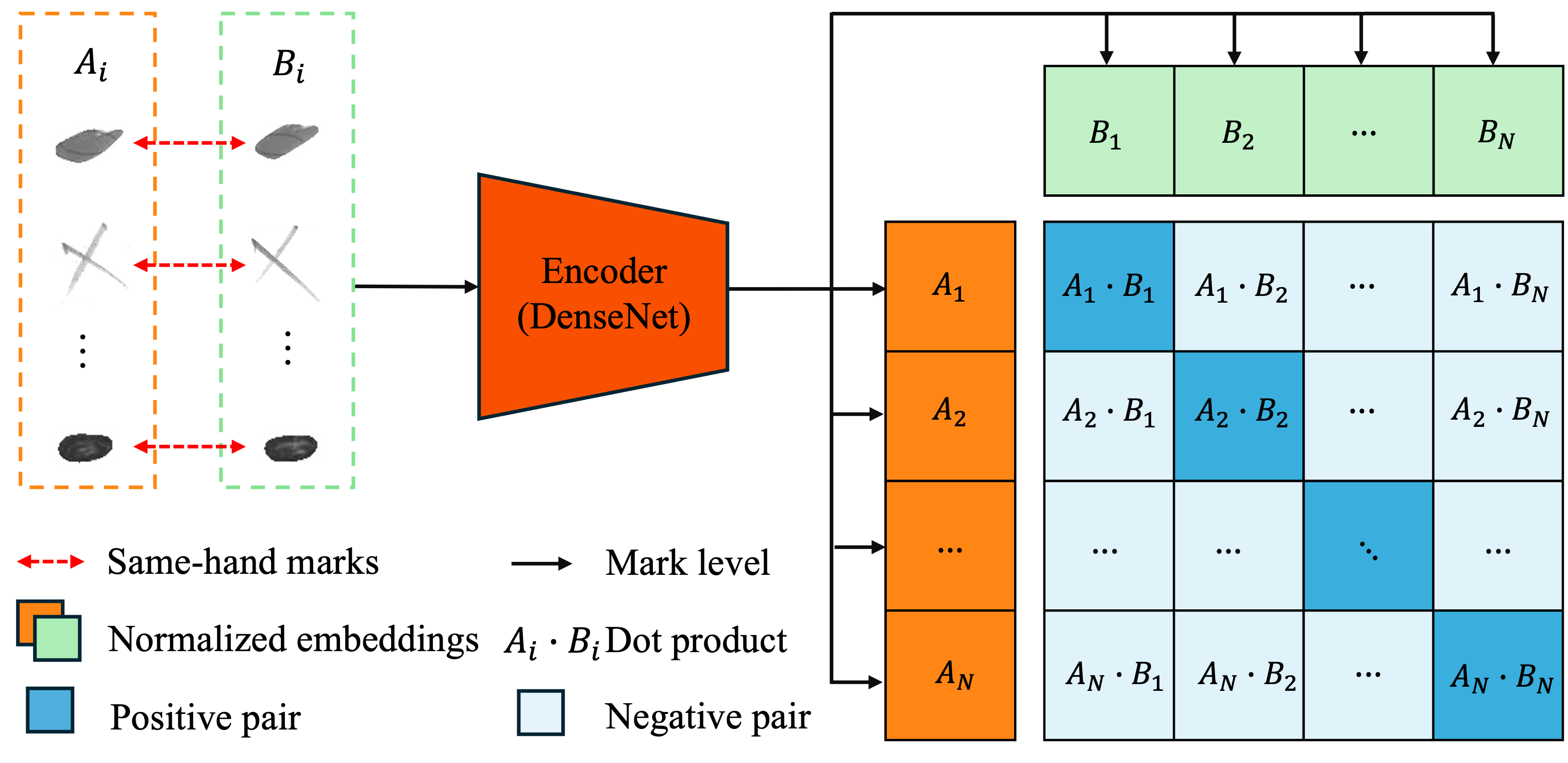}

    \caption{{The proposed contrastive learning model (detailed in Section \ref{sec:model})} }

    \vspace{-0.4cm}

    ~\label{fig:fig_1}

  \end{figure}
\vspace{-0.4cm}

\section{The Model Design}
\label{sec:model}

\begin{figure}[ht]
    \centering

    \includegraphics[width=0.45\textwidth]{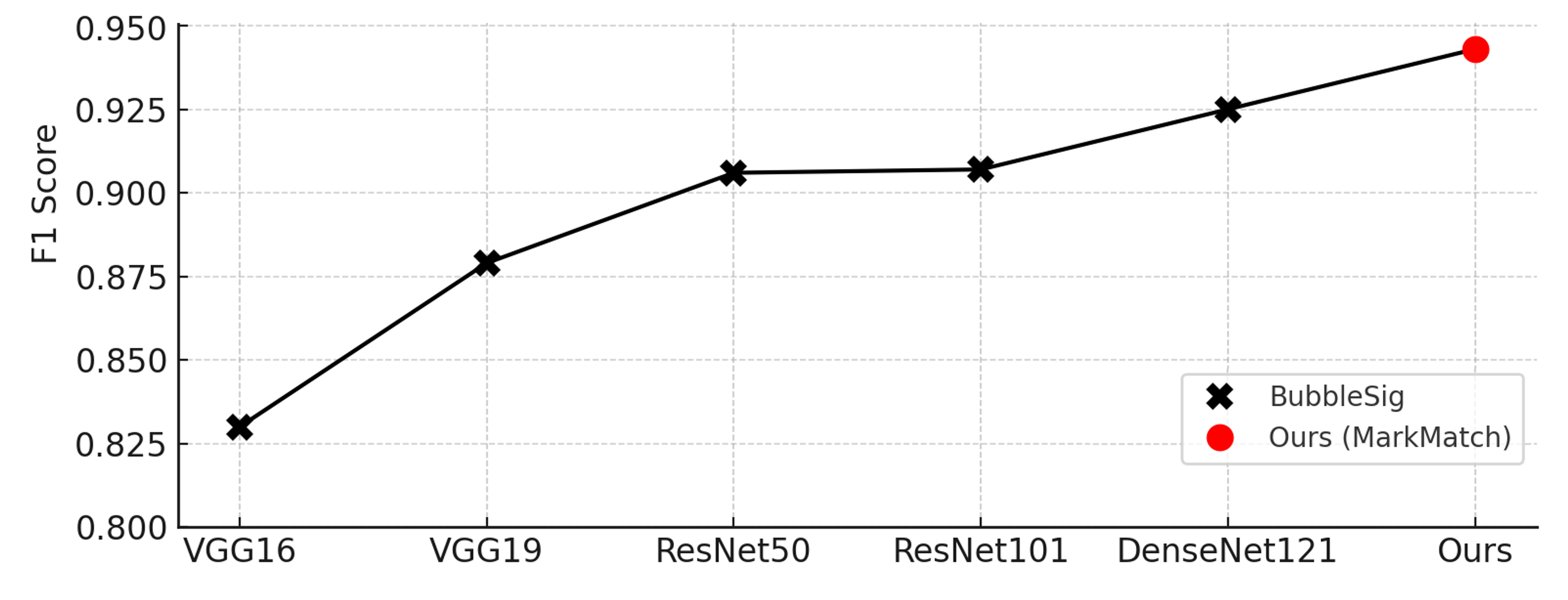}
    \vspace{-0.3cm}

\caption{F1 comparison between \textit{BubbleSig}~\cite{zhao2024bubblesig} variants and our model}
    \vspace{-0.6cm}

    ~\label{fig:fig_3}

  \end{figure}

As illustrated in Fig.~\ref{fig:fig_1}, we propose a contrastive learning model trained to distinguish positive pairs from a large set of negatives in each batch. Each ballot mark image is independently passed through a shared DenseNet121 encoder to obtain a fixed-length embedding. A training batch is constructed by sampling two aligned sets of ballot marks, {\small${A_i}$} and {\small${B_j}$}, for {\small $i, j = 1, \dots, n$}, such that each {\small$(A_i, B_i)$ }corresponds to the same writer and is a positive pair, while each {\small$(A_i, B_j, i \ne j)$ } is a negative pair. Let {\small $f(\cdot)$} denote the shared encoder. All embeddings are L2-normalized, and a similarity matrix {\small $S \in \mathbb{R}^{n \times n}$} is computed using scaled dot product: {\small $S_{ij} = \langle f(A_i), f(B_j) \rangle / \tau$}, where {\small$\tau$} is a temperature scaling factor set as 0.07. We collect softmax-based cross-entropy losses in both row and column directions. This encourages each embedding to align closely with its true positive counterpart while minimizing similarity to other negatives within the batch. We further apply binary cross-entropy to sigmoid-activated diagonal scores, enhancing the model’s certainty on positive pairs. The loss is defined as:

{\small
\[
\mathcal{L} = \frac{1}{2} \left( \mathcal{L}_{\text{CE}}^{\text{row}} + \mathcal{L}_{\text{CE}}^{\text{col}} \right) + \alpha \cdot \mathcal{L}_{\text{BCE}}
\]
}

where {\small $\mathcal{L}_{\text{CE}}^{\text{row}}$} denotes the row-wise cross-entropy (CE) loss computed as {\small $\frac{1}{n} \sum_{i=1}^{n} \text{CE}(\text{softmax}(S_{i,:}), Y_{i,:})$}, and {\small $\mathcal{L}_{\text{CE}}^{\text{col}}$} is its column-wise counterpart {\small $\frac{1}{n} \sum_{j=1}^{n} \text{CE}(\text{softmax}(S_{:,j}), Y_{:,j})$}. The ground truth matrix {\small \( Y \in \{0,1\}^{n \times n} \)} assigns label 1 to diagonal entries (positive pairs) and 0 to all off-diagonal entries (negative pairs). {\small $\mathcal{L}_{\text{BCE}}$} represents the binary cross-entropy (BCE) loss applied to the diagonal entries: {\small $\frac{1}{n} \sum_{i=1}^{n} \text{BCE}(\sigma(S_{ii}), Y_{ii})$}, where {\small $\sigma(\cdot)$} is the sigmoid function. The coefficient {\small $\alpha$} is set to 1 based on empirical tuning. We train and evaluate our model using the same large-scale dataset and the configuration as \textit{BubbleSig}~\cite{zhao2024bubblesig}. As shown in Fig.~\ref{fig:fig_3}, our model achieves an F1 score of 0.943.

    \vspace{-0.2cm}

For deployment, \textit{MarkMatch} adopts a retrieval-based formulation. Given a query mark, the model computes similarity scores against a pool of candidates and applies a softmax over these scores to obtain a normalized ranking. The top-ranked marks are returned as the most likely same-hand instances. This formulation aligns naturally with contrastive learning and yields more reliable predictions than binary classification on isolated pairs, particularly in visually ambiguous cases.


\section{Demonstration Features}\label{sec:sys}

\begin{figure}[ht]
    \centering
    \includegraphics[width=0.43\textwidth]{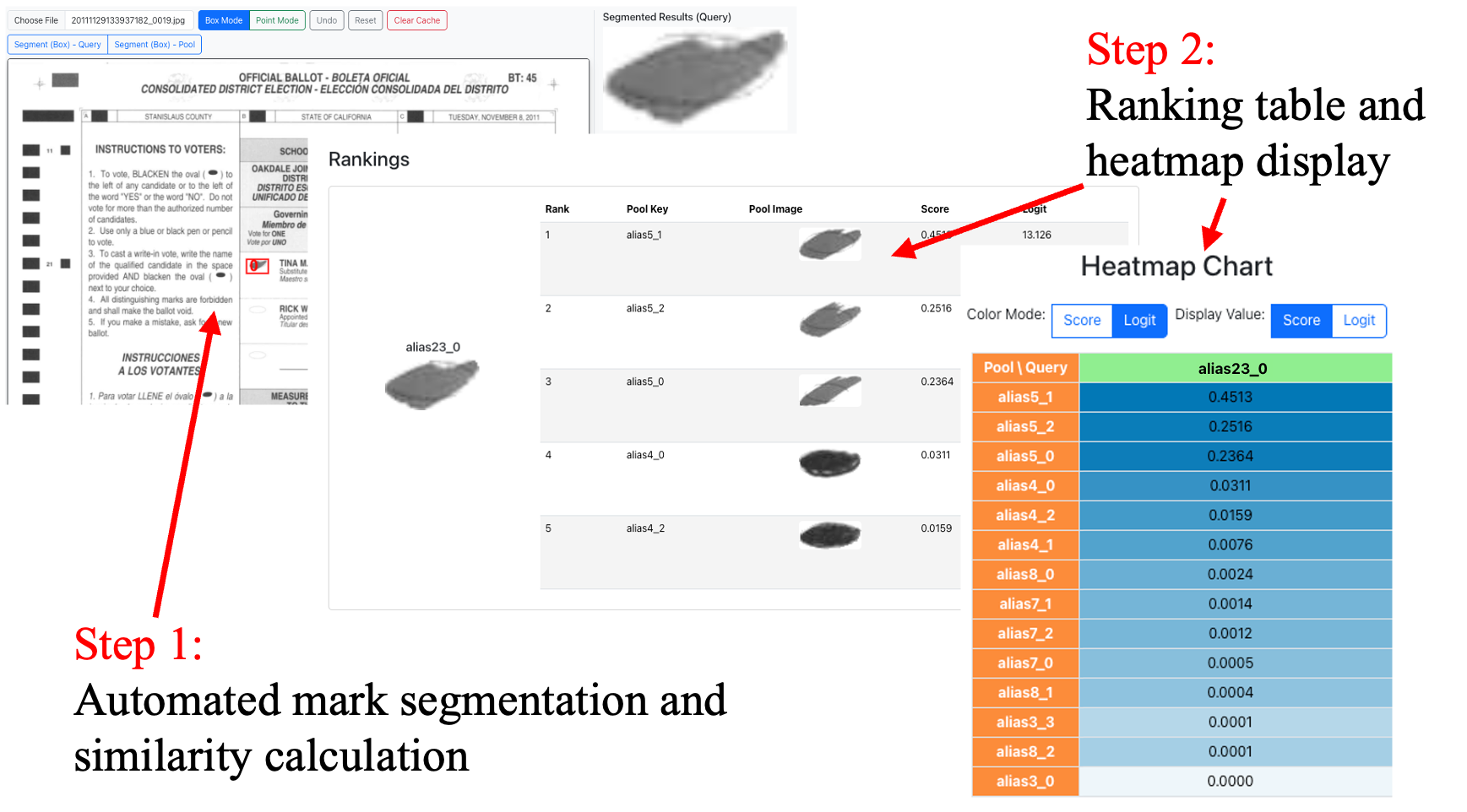}
        \vspace{-0.3cm}


\caption{\textit{MarkMatch} interface. The system segments marks using prompt-based inputs (box or point), computes similarity between the query (green) and candidates (orange), and visualizes softmax-normalized scores via heatmap and ranking. To preserve privacy, marks are anonymized using letter-number aliases (e.g., \texttt{alias5\_0}). In this example, three top-ranked matches from the same ballot (\texttt{alias5}) are retrieved for the query mark \texttt{alias23\_0}, and manual review confirms both ballots were filled by the same individual.}

    ~\label{fig:fig_last}

  \end{figure}

\textit{MarkMatch} provides an interactive platform for analyzing and comparing hand-drawn ballot marks through visual similarity. The system begins by using SAM to extract individual mark segments from a scanned ballot image (Step 1 in Fig. \ref{fig:fig_last}). Users can interactively select regions using bounding boxes or point clicks, allowing online and flexible segmentation across varying mark shapes and ballot layouts. Once segmented, each selected mark can serve as a query, which is compared against a pool of candidate marks. For every query, the system computes a similarity score with every mark in the pool using our contrastive model. These results are visualized through two components, shown as Step 2 in Fig. \ref{fig:fig_last}:

\begin{itemize}[leftmargin=*]

\item \textbf{Ranking Table}: For each query mark, the system displays the top-5 retrieved pool marks ranked by softmax-normalized similarity scores. For each match, the table includes the candidate’s rank, filename, image, softmax score, and raw similarity logit. This view allows auditors to assess both the statistical evidence and visual resemblance behind each prediction.

\item \textbf{Heatmap Visualization}: A matrix showing softmax-normalized similarity scores between candidate pool marks (rows) and query marks (columns). Each cell reflects the model’s confidence that a pool mark matches the corresponding query. This global view reveals retrieval confidence distributions across all pairs.
\end{itemize}

    \vspace{-0.3cm}

\textbf{Demo Process}:  Attendees will hand-mark multiple paper ballots, which demo assistants will photograph and upload to MarkMatch. The system segments each mark using SAM, extracts feature embeddings, and computes pairwise similarity scores against a candidate pool. Top-k most similar marks are returned and visualized through ranked tables and heatmaps.

\section{Conclusions \& Acknowledgement}

\label{sec:conclusions}

We presented \textit{MarkMatch}, a retrieval-based system for detecting same-hand ballot marks using contrastive learning. By modeling relative similarity across batches, it improves robustness in ambiguous and noisy cases. \textit{MarkMatch} integrates SAM for prompt-based segmentation on the fly and provides intuitive visualizations via heatmaps and ranked tables, offering election auditors an efficient, non-biometric tool for scalable, transparent ballot review. This work was supported by NSF CNS-2154589 and 2154507, “Collaborative Research: SaTC: CORE: Medium: Bubble Aid: Assistive AI to Improve the Robustness and Security of Reading Hand-Marked Ballots,” \$1,200,000, 10/01/2022-09/30/2026.


\bibliographystyle{IEEEbib}


\bibliography{IEEEabrv,icme2023_ref_file}

\end{document}